\pdfoutput=1

\documentclass[11pt]{article}

\usepackage[]{ACL2023}

\usepackage{times}
\usepackage{latexsym}
\usepackage[linewidth=1pt]{mdframed}
\usepackage{hyperref}
\usepackage{color}
\usepackage{booktabs}
\usepackage{amssymb,stackengine,graphicx}
\usepackage{multirow}
\usepackage{textpos}
\definecolor{mygray}{gray}{0.9}
\usepackage{comment}

\usepackage{xcolor}

\usepackage{pifont}
\usepackage[T1]{fontenc}

\usepackage[utf8]{inputenc}
\usepackage{adjustbox}
\usepackage{microtype}

\usepackage{inconsolata}

\newcommand{\cmark}{\ding{51}}%
\newcommand{\xmark}{\ding{55}}%
\newcommand{\gray}[1]{\textcolor{gray}{#1}}
\title{Advancing Large Language Models to Capture Varied Speaking Styles and Respond Properly in Spoken Conversations}

\author{Guan-Ting Lin, Cheng-Han Chiang, Hung-yi Lee\\
  National Taiwan University, \\
  Taiwan \\
  \texttt{\{f10942104, hungyilee\}@ntu.edu.tw}, \texttt{dcml0714@gmail.com} \\}

\begin{document}
\maketitle
\begin{abstract}
In spoken dialogue, even if two current turns are the same sentence, their responses might still differ when they are spoken in different styles.
The spoken styles, containing paralinguistic and prosodic information, mark the most significant difference between text and speech modality. 
When using text-only LLMs to model spoken dialogue, text-only LLMs cannot give different responses based on the speaking style of the current turn.
In this paper, we focus on enabling LLMs to \textit{listen to} the speaking styles and respond properly.
Our goal is to teach the LLM that "\textit{even if the sentences are identical if they are spoken in different styles, their corresponding responses might be different}".
Since there is no suitable dataset for achieving this goal, we collect a speech-to-speech dataset, \textbf{StyleTalk}, with the following desired characteristics: when two current speeches have the same content but are spoken in different styles, their responses will be different.
To teach LLMs to understand and respond properly to the speaking styles, we propose the \textbf{Spoken-LLM} framework that can model the linguistic content and the speaking styles.
We train Spoken-LLM using the StyleTalk dataset and devise a two-stage training pipeline to help the Spoken-LLM better learn the speaking styles.
Based on extensive experiments, we show that Spoken-LLM outperforms text-only baselines and prior speech LLMs methods.
\footnote{Demo of the StyleTalk dataset and output of Spoken-LLM are at \url{https://sites.google.com/view/spoken-llm/home}. Code and dataset are available at \url{https://github.com/DanielLin94144/StyleTalk}.}

\end{abstract}

\section{Introduction}
 

Large Language Models (LLMs) have demonstrated remarkable capabilities in dialogue generation, natural language understanding, and commonsense reasoning~\citep{llm1, gpt4}. While LLMs mostly focus on text modality, speech represents the most natural form of human communication in our daily lives. In this work, we aim to inject speech modality for modeling \textit{spoken conversation} with Multi-modal LLMs (MM-LLMs). The main goal is to develop a humanizing agent capable of listening, understanding, and engaging in dialogue with humans, ultimately leading to higher user satisfaction. 

\begin{figure}[t]{}
    \centering
\includegraphics[width=1\linewidth]{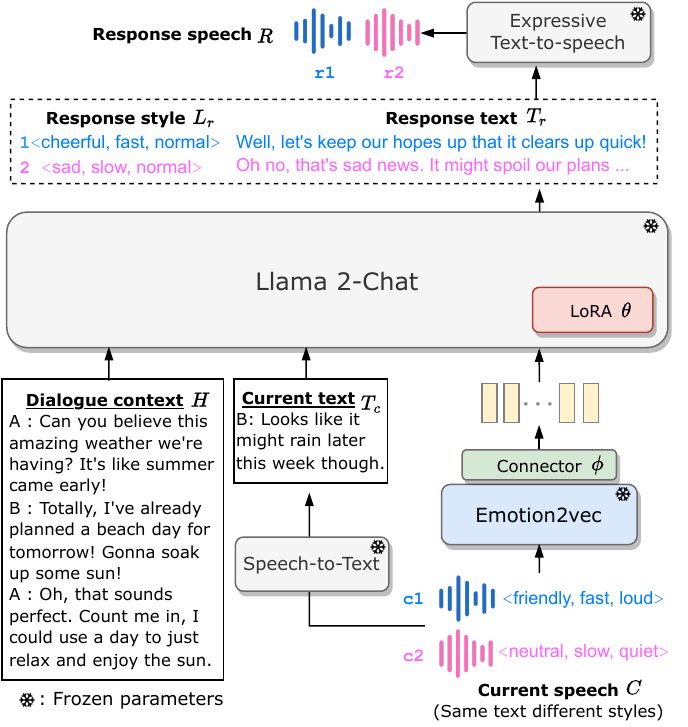}
    \caption{The overview framework of Spoken-LLM. (\texttt{c1},\texttt{r1}) and (\texttt{c2},\texttt{r2}) are the current and response speech sample pairs. \texttt{c1} and \texttt{c2} are fed into the model individually.}
    \label{fig:framework}
\end{figure}

Speech signals contain linguistic aspects (words, phonetics, syntax, and semantics), paralinguistic elements (emotions and speaker characteristics), and prosodic factors (speaking style, emphasis, and attitude). In human conversation, while the dialogue primarily relies on the lexical aspect, the speaking styles convey rich information beyond text, and can even alter the semantics of the spoken sentences~\citep{mustard}. Neglecting spoken styles can lead to misinterpretation of communication or unnatural human interaction. For example, as shown in Figure \ref{fig:framework}, the current speech with the same current text (\textit{Looks like it might rain later this week though.}) but different speaking styles. The friendly speaking style leads to a cheerful response while speaking in a slow and neutral tone leans toward a sad and negative response. 


Although there are recent studies on MM-LLMs for speech/audio and text, most of the existing studies focus on \textit{content-centric} Spoken Language Modeling (SLM)~\cite{gslm, pgslm}, joint text and speech processing tasks~\citep{audiopalm, toward_joint, voxtlm, spectron, speechgpt} or general audio perception and hearing ability~\citep{salmonn, ltu-as, pengi}. There is less attention on spoken dialogue with advanced methods and suitable datasets for modeling paralinguistics and speaking styles of spoken responses.

To model spoken dialogue with a generative language model, dGSLM~\citep{dgslm} proposes a dual-tower SLM on discrete speech units to model two-channel spoken dialogue, but the generated spoken sentences lack semantic meaning. ParalinGPT~\citep{paralingpt} organizes tasks in the sequence of current paralinguistic attribute prediction, response paralinguistic attribute prediction, and response text generation with autoregressive conditioning. However, it only uses the speech sentiment as speaking style, which might be primarily based on textual information, and how the speaking styles affect the spoken response is unclear. A concurrent work E-chat~\citep{echat} enhances LLM to generate responses in different emotional contexts, but the training and evaluation data are entirely generated by GPT-3.5 without human supervision, equivalent to distillation and prompting of GPT-3.5. It can only generate response text, constraining its capacity to control response style or speech-to-speech modeling.

To overcome the current limitation, we collect a novel speech-to-speech conversational dataset named \textbf{StyleTalk}. This dataset is the first spoken conversation benchmark with \textit{the same dialogue context and input sentence in different speaking styles, accompanied by corresponding expressive spoken responses} for speech-to-speech modeling. The dataset will be released upon the paper's acceptance.



Based upon the StyleTalk dataset, we propose a multi-modal two-stage training method named \textbf{Spoken-LLM} for spoken dialogue modeling. Spoken-LLM is a fusion of the widely-used open-sourced LLM (Llama 2-Chat~\citep{llama2}) and a self-supervised speech emotion representation model (emotion2vec~\citep{emotion2vec}). The proposed model can predict response speaking style and text, enabling the subsequent expressive Text-to-Speech (TTS) model to generate natural and diverse speech responses. We validate the performance through objective and subjective evaluations of spoken responses. With the same backbone model, the proposed method outperforms the text and speech LLM baseline in lexical/semantic similarity and response style F1 score. The human evaluation also indicates that the proposed method yields more reasonable and proper response speech than the text-only LLM baseline approach.

\begin{table}[t]
\centering
\adjustbox{width=0.5\textwidth}{
\begin{tabular}{ccccc}
\toprule
\textbf{Dataset} & \textbf{S2S} & \textbf{Expressive}                                  & \textbf{Purpose}    & \textbf{Diff styles\&resp} \\ \hline
IEMOCAP          & \cmark        & \cmark                                         & recognition         & \xmark                             \\ 
Switchboard          & \cmark        & \cmark                              & recognition         & \xmark            \\
MUStARD          & \cmark        & \cmark                                         & recognition         & \xmark                             \\
SEMAINE          & \cmark        & \cmark                              & recognition         & \xmark            \\
MELD             & \cmark        & \cmark                              & recognition         & \xmark                             \\
MEISD            & \cmark        & \cmark                   & recognition         & \xmark             
\\
MSP-improv            & \cmark        & \cmark                   & recognition         & \xmark             
\\
SCQA            & \xmark        & \xmark                                              & question answering  & \xmark  \\ 
NMSQA            & \cmark        & \xmark                                               & question answering  & \xmark                             \\
OpenSAQA$^*$         & \xmark        & \cmark & question answering  & \xmark                             \\
E-chat200$^*$           & \xmark        & \cmark                                         & dialogue generation  & \xmark                             \\ \hline
\textbf{StyleTalk}      & \cmark        & \cmark                       & dialogue generation & \cmark  \\ \bottomrule
\end{tabular}
}
\caption{The list of spoken conversation datasets. ``S2S" means speech-to-speech, and ``Diff styles\&resp" stands for \textit{the same sentence in different speaking styles and responses}. In the ``Purpose" column, ``recognition" refers to recognizing the speaking style attributes in the speech, ``question answering" means the task is formulated as the (question, answer) pair, and ``dialogue generation" is the general chatbot agent to response any kinds of input. The datasets noted with $^*$ are purely generated by LLM. }
\label{tab:dataset}
\end{table}



\section{Dataset: StyleTalk}

\subsection{Overview}
StyleTalk is a speech-to-speech conversation dataset.
Each sample in the dataset comprises dialogue context (in text), current turn in speech (annotated with speaking style), and the response turn in speech (annotated with speaking style) (illustrated in Figure \ref{fig:framework}).

StyleTalk features the following characteristics:
Multiple samples in StyleTalk share the same dialogue context, the text of the current input turn, but they have different responses speech since the speaking style of the current turn is different.
To the best of our knowledge, no existing corpora focus on such a characteristic.

By training on this dataset, we hope the LLM can learn to use the dialogue context and current turn, specifically, the speaking style, to predict the next turn.
Given that speaking styles convey additional information beyond text, incorporating style modeling helps to disambiguate human intent and facilitates dialogue engagement.

\begin{figure}[t]{}
    \centering
\includegraphics[width=0.95\linewidth]{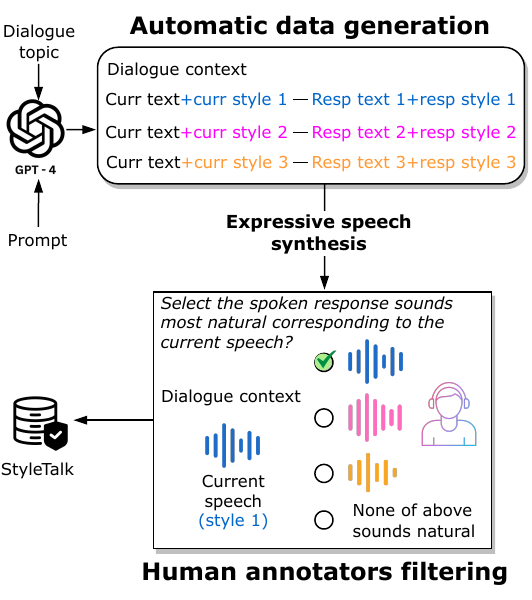}
    \caption{Data collection pipeline of StyleTalk. The details of instruction and prompt template are in the Appendix.}
    \label{fig:data}
\end{figure}

\subsection{Data Collection}
In this section, we introduce how the dataset is collected.
The data collection pipeline includes three stages: (1) using LLM for data generation text dialogue with style annotation, (2) using an expressive TTS model to synthesize speech from text dialogue, and (3) recruiting human annotators to filter the dataset. 
We illustrate the data collection pipeline in Figure \ref{fig:data}.

\subsubsection{LLM for Data Generation}
Crafting a scenario with the same context and words but expressed in different speaking styles is a non-trivial task. Most dialogue corpora typically consist of one style, making it challenging to study the impact of various speaking styles on spoken responses. 

Recently, LLMs have demonstrated human-level knowledge and powerful data generation capabilities when provided with well-designed prompts and instructions. In light of this, we propose leveraging GPT-4~\citep{gpt4} to generate spoken \textbf{dialogue set} consisting of a dialogue context, the same sentence presented in three different speaking styles, and three corresponding responses. To let LLM understand the speaking style information, the speaking style is represented in text by surrounded by special marker, for example, <\textit{emotion, speed, volume}>. 
To increase the diversity of the dialogue, we prompt the GPT-4 with 17 common daily dialogue topics: 
school, work, family, health, entertainment, travel, food, sports, finance, technology, music, movies, books, games, beauty, shopping, and weather.
Additionally, we use decoding with temperature sampling to ensure diversity in the dataset.
The prompt template is shown in the appendix \ref{fig:prompting}. 

\subsubsection{Expressive Speech Synthesis}
To generate high-quality speech with style and prosody control, we utilize an \href{https://azure.microsoft.com/en-us/products/ai-services/text-to-speech}{industrial-grade Microsoft Azure Text-to-Speech (TTS) system}\footnote{https://azure.microsoft.com/en-us/products/ai-services/text-to-speech}. For the speaking style, we employ \textbf{emotion} (neutral, cheerful, sad, friendly, unfriendly), \textbf{speeds} (slow, medium, fast), and \textbf{volumes} (quiet, medium, loud) for prosodic control. There are nine speakers, with four male and five female speakers.

\subsubsection{Human Annotator Filtering}
While LLMs can effectively follow instructions and generate reasonably coherent dialogue samples, LLMs are trained on textual data and lack exposure to human-human spoken dialogue. Additionally, the expressive TTS system may not achieve perfect naturalness and style-following in synthesizing speech under style conditions. The automatically generated data may exhibit unnatural characteristics for human speakers. Therefore, additional examination is necessary to check the quality of the speech data and the overall naturalness of spoken dialogue sample pairs.

To ensure data quality, we request human listeners to participate in a listening test conducted on the Amazon Mechanical Turk platform. An illustration of the listening test is provided in Figure \ref{fig:data}. In this evaluation, participants are presented with a dialogue context text, the current spoken turn and three response spoken turns. They are then instructed to choose the most suitable response among the three options. Alternatively, if they perceive all three responses as unnatural, they can select the option "None of the above is natural." Participants need to be aware of the style of the current turn and \textit{differentiate between the three response turns to identify the most natural one}. Through this evaluation, we aim to filter out sample pairs that are deemed unnatural or indistinguishable. Details are shown in Appendix \ref{sec:manual}. We found out that only around 33\% samples successfully passed the human filtering process. This suggests that LLM-generated spoken samples are either \textit{not natural to human perception} or \textit{the speaking style does not distinctly influence spoken responses}. 


\subsection{Data split}
 After manual filtering, we split the filtered data into training and evaluation sets with dialogue sets. ``Sample" means a current and response speech pair. The detailed data statistics are shown in Appendix Table \ref{tab:data-statistic}.\\
\textbf{Train set}: The training set is carefully curated through manual filtering, resulting in 1,878 dialogue sets and 1,986 samples. \\
\textbf{Evaluation set}: The evaluation set contains 486 dialogue sets and 981 samples, most of the dialogue sets have two to three speaking styles for the current text. \\
In addition to the train and evaluation set, a fully LLM-generated \textbf{unfiltered set} is introduced for data augmentation since the size of the training set is limited. The unfiltered set consists of 5,777 dialogue sets and 16,472 samples. It is crucial to note that this data is not subject to human supervision, and as such, the samples may not align perfectly with human standards. 

\section{Spoken-LLM framework}

\subsection{Overview}
The framework of Spoken-LLM is illustrated in Figure \ref{fig:framework}. The main components include the large language models, speaking style encoder, speech-to-text conversion, and expressive TTS system. $D_s$ and $D_t$ denote the dimension of the speech encoder's output and LLM's input space, respectively.

We formulate the task as follows: given a multi-turn spoken dialog with dialogue context $H$ in text $T_{h}$, a current turn $C$ comprising speech $S_{c}$ and text $T_{c}$. The prediction response speech $R$ includes response style $L_{r}$ and response text $T_{r}$. Note that we use the ground truth transcripts $T_c$ of the current turn, since addressing speech recognition errors is not the focus of this work. The discussion of using ASR prediction is in section \ref{sec:asr}. 

\subsection{Large Language Model}
This study adopts the open-sourced Llama 2-Chat 7B model, derived from the fine-tuned version of Llama 2~\citep{llama2}, exhibiting optimized dialogue generation capabilities. Throughout the training process, the Llama 2-Chat model remains frozen, and we introduce the trainable LoRA adapter~\citep{lora} for parameter-efficient fine-tuning.

\subsection{Speech Style Encoder}

Among the self-supervised speech models~\citep{superb, superb-prosody}, emotion2vec~\citep{emotion2vec} achieves state-of-the-art performance on diverse paralinguistics-related tasks. Precisely, it extends the data2vec 2.0~\citep{data2vec2} with both utterance-level and frame-level loss using emotional speech data, and extra chunk token embeddings are used to capture utterance-wise information.

We choose emotion2vec as the speech encoder to extract universal paralinguistic and prosody embeddings. Two approaches are used for feature extraction. (1) \textit{Utterance-level averaging embedding (\textit{utt})}: which involves a simple averaging of frame-wise representations to create an utterance-level embedding. The embedding is in $1 \times D_s$ dimension. (2) \textit{Chunk embedding}: emotion2vec learns 10 extra chunk token embeddings to capture both fine-grained and global speech information. Chunk embeddings are in $10 \times D_s$ dimension.

 A lightweight \textit{Connector} module with layer normalization and a linear model is utilized to project the speech embeddings into the dimension of the language model's input space (from $D_s$ to $D_t$). Only the parameters of the connector are updated, while the emotion2vec model remains frozen. The number of trainable parameters for utterance and chunk embeddings are the same.

\subsection{Spoken Dialogue Modeling}
\textbf{1st-stage: style alignment}: The first-stage training is used to align the speech embedding with LLM input space. To achieve this, the frozen LLM has to predict the current input style. Only the connector $\phi$ is trained. The training objective is to minimize the cross-entropy loss for classifying $L_c$: 
\begin{eqnarray}
\mathcal{P}(L_c|C, I_1; \phi), 
\end{eqnarray}
where $C = \{T_{c}, S_{c}\}$. $I_1$ is the task instruction shown in Appendix \ref{sec:instruction}. Since this training stage requires a reasonable amount of data to have better alignment, we use the current speech from the unfiltered set for training.\\

\noindent{\textbf{2nd-stage: spoken response modeling}}: After the LLM can understand the speech embedding, the LLM is optimized to predict the response style and response text by training the LoRA adapter $\theta$ and speech connector $\phi$. The second-stage training objective is the causal language modeling cross-entropy loss to predict the response style $L_r$ then response text $T_r$: 

\begin{small}
\begin{eqnarray}
    \mathcal{P}(R|H, C, I_2; \theta, \phi) &=& \mathcal{P}(L_{r}|H, C, I_2; \theta, \phi) \nonumber \\
    && \mathcal{P}(T_{r}|H, C, L_{r}, I_2; \theta, \phi)
\end{eqnarray}
\end{small}
where $H = T_{h}$ and $I_2$ is the task instruction shown in Appendix \ref{sec:instruction}. The speaking style label $L_r$ is integrated into the text through special bracket markers with the format <\textit{emotion, speed, volume}>. 
$T_h$ and $T_c$ are fed into LLM subword embedding, and $S_c$ is passed through the speech encoder plus the connector. We concatenate the resulting continuous embeddings as the input prompt for LLM.

\begin{table*}[]
\centering
\adjustbox{width=0.85\textwidth}{
\begin{tabular}{l|cccc|ccc}
\toprule
\multirow{2}{*}{\textbf{Method}}           & \multicolumn{4}{c}{\textbf{Response text}}                 & \multicolumn{3}{|c}{\textbf{Response style}} \\
                                   & \textbf{BLEU} & \textbf{ROUGE$_l$} & \textbf{METEOR} & \textbf{BERT$_{f1}$}  & \textbf{F1$_{emotion}$} & \textbf{F1$_{speed}$} & \textbf{F1$_{volume}$}\\ \hline
                                  
Text-LLM (text-only)    &  3.1    &  16.2     &   17.4     &   75.3   &    17.5       &     37.1      &    41.9       \\
Text-LLM (cascaded)                     &   3.2   &   17.3   &       19.1              &  76.0    &     37.5       &    52.9       &   65.6        \\ 
\gray{Text-LLM (upper bound)}                       &  \gray{4.0}    &   \gray{17.9}    &  \gray{19.6}      &  \gray{76.3}    &   \gray{40.2}         &    \gray{53.5}       &   \gray{65.8}        \\ \hline
ParalinGPT-\textit{utt}   & 3.1  & 16.8  & 18.5   & 75.9   &    32.3        &   51.9       & 64.8          \\
ParalinGPT-\textit{chunk}        &  3.1    &      16.5 & 18.2 &  75.8    &    34.0        &   54.8        & \textbf{65.8}          \\ \hline
Spoken-LLM-\textit{utt}  &  2.8   &   16.6    & \textbf{20.2}  & 75.8&  47.4   &  61.5       &   56.5   \\ 
Spoken-LLM-\textit{chunk}  &  \textbf{4.0}    &  \textbf{17.8}     &   19.4 &  \textbf{76.3}   &    \textbf{49.6}       &     \textbf{62.1}      & 61.1 \\ 
\bottomrule        
\end{tabular}
}
\caption{Main results comparing text-LLM, ParalinGPT, and Spoken-LLM. \textit{utt} and \textit{chunk} refer to utterance-wise and chunk-wise speech embedding from emotion2vec. The Text-LLM (upper bound) method is the cascaded text-LLM with ground truth style labels.}
\label{tab:main_result}
\end{table*}

\noindent{\textbf{Warmup pre-training}}: Given the limited size of the human-annotated training set, we propose leveraging the unfiltered set for model warmup pre-training. This allows the model to grasp general knowledge and understand the structure of the dialogue modeling task. Subsequently, we fine-tune the model on the training set to align with human perception, utilizing a smaller learning rate for stable training. This warmup training strategy is designed to mitigate overfitting on the small training data while maintaining good performance. 


\subsection{Inference}
Once the model has completed training, when presented with a dialogue context and current speech input, the initial step involves converting the speech into text through either ground truth text in the oracle setup or an Automatic Speech Recognition (ASR) model in the ASR setup. Then, the Spoken-LLM generates response style and response text sequentially. The representation of the response style is surrounded by special bracket tokens, designed to enable the decoding of both the response style and text. Leveraging the capability of an expressive TTS model to control the response speaking style, we can synthesize the generated response back into speech. This synthesis takes into account the identified response style, and the generated response text, resulting in a synthesized speech output that is not only coherent but also aligns with the desired style and content.

\section{Experiments}

\subsection{Baseline method}
All baseline methods are fine-tuned on the same amount of training data and warmup pre-training, with the identical LLM backbone and speech style encoder for ParalinGPT. \\
\textbf{Text-LLM (text-only)}: The initial simple baseline is built by simply fine-tuning text-to-text LLM on StyleTalk. This serves as a performance reference to evaluate the model's capability without knowing any explicit speaking style information. Since the model cannot predict the response style, A randomly selected response style is assigned for this method to synthesize expressive speech.\\
\textbf{Text-LLM (cascaded)}: One can represent the style information in text to enable the model to better predict the response style and text. This approach, referred to as the cascaded pipeline method, involves cascading a style recognition model\footnote{We use the Spoken-LLM-\textit{chunk} 1st-stage model as the style recognition model, which achieves 86.8, 99.2, 64.0 f1 scores on current emotion, speed, volume prediction, respectively.} with the text LLM. The Text-LLM (upper bound) method is the cascaded text-LLM with ground truth style labels. \\
\textbf{ParalinGPT~\citep{paralingpt}}: The \textit{serialized multitasking} approach proposed by ParalinGPT is a sequential conditioning mechanism, unifying current style prediction, response style prediction, and response text generation within an auto-regressive chain. The main difference between ParalinGPT and Spoken-LLM is that Spoken-LLM performs two-stage training (style alignment for current speech then focus on the response speech), but ParalinGPT directly models them in an auto-regressive chain, which might be prone to error propagation if the incorrect current style prediction or focusing too much on the current style. 


\subsection{Evaluation Metrics}
\textbf{Objective evaluation}: For automatic evaluation of response text, we adopt the widely-used text generation metric, including lexical-level score (BLEU~\citep{bleu}, ROUGE~\citep{rouge}, METEOR~\citep{meteor}), and semantic-level (BERT Score~\citep{bertscore})\footnote{Score calculated by \href{https://huggingface.co/docs/evaluate/}{Hugging face Evaluate package}}. For response style evaluation, since the style attributes are categorical,  we calculate the Weighted F1 score for speaking emotion, speed, and volume. \\
\textbf{Subjective evaluation}: We perform the human evaluation on a set of 200 samples using an A/B test for model comparison. Three human evaluators are assigned to each sample, and they are instructed to rate the model based on both the generated text and speech. The details of subjection evaluation are in Appendix \ref{sec:subjective-eval}.

\begin{figure}[t]{}
\centering
\includegraphics[width=0.85\linewidth]{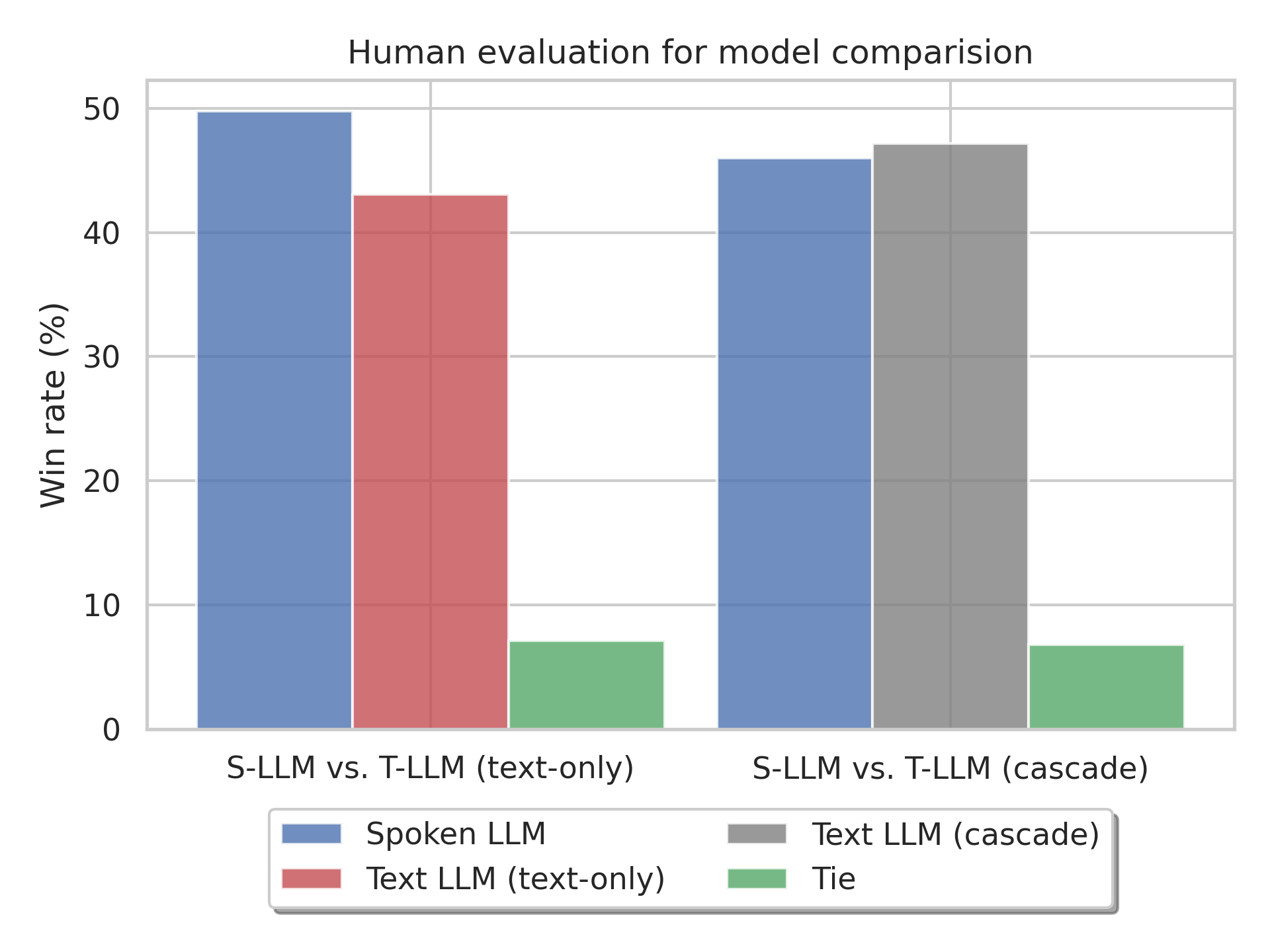}
    \caption{Human evaluation result comparing Spoken-LLM-chunk with Text-LLM (text-only) and Text-LLM (cascaded).}
    \label{fig:human_eval}
\end{figure}

\begin{table*}[]
\centering
\adjustbox{width=0.85\textwidth}{
\begin{tabular}{c|cccc|ccc}
\toprule
\multirow{2}{*}{\textbf{Training data}} & \multicolumn{4}{c}{\textbf{Response text}}   & \multicolumn{3}{|c}{\textbf{Response style}} \\
                                & \textbf{BLEU} & \textbf{ROUGE$_l$} & \textbf{METEOR} & \textbf{BERT$_{f1}$}       & \textbf{F1$_{emotion}$} & \textbf{F1$_{speed}$} & \textbf{F1$_{volume}$}     \\ \hline
train                         &  2.9 & 15.7  & 17.8  & 75.5   & 45.5      & 61.6     & \textbf{61.7}      \\
unfiltered                  & 3.4 & 17.0 & 18.7 & 75.7   &  44.1     &   61.2    &   56.5    \\
unfiltered$\rightarrow$train               & \textbf{4.0}    &  \textbf{17.8}     &   19.4 &  \textbf{76.3}   &    \textbf{49.6}       &     \textbf{62.1}      & 61.1 \\ \bottomrule    
\end{tabular}
}
\caption{Different training strategy and data usages on Spoken LLM-\textit{chunk} method. ``$\rightarrow$" indicates the two-stage warmup training pipeline.
}
\label{tab:warmup}
\end{table*}

\begin{table}[]
\centering
\adjustbox{width=0.35\textwidth}{
\begin{tabular}{ll}
\toprule
\textbf{Method} & \textbf{self-BLEU} \\ \hline
\gray{Ground truth}         & \gray{8.2}       \\
Text-LLM (text-only) & 100.0      \\
Text-LLM (cascaded)   & 11.2      \\
ParalinGPT-\textit{chunk}   & 11.3      \\
Spoken-LLM-\textit{chunk}           & \textbf{10.9}     \\ \bottomrule
\end{tabular}
}
\caption{The dialogue set-level self-BLEU score for different methods on the evaluation set.}
\label{tab:self-bleu}
\end{table}

\subsection{Main Results} 
\textbf{Spoken-LLM outperforms speech and text baselines}: Table \ref{tab:main_result} shows the result on objective evaluation. Firstly, for the text-LLM baseline on response text metrics, we observe that adding current speech style information yields significantly better performance than the text-only method, indicating that recognizing the style information is beneficial to predict textual response. Next, we compare the Text-LLM and speech ParlinGPT baseline. ParalinGPT consistently outperforms the Text-LLM method on the response text metrics. However, on the response style, the text-LLM (cascaded) is better than ParalinGPT-\textit{utt}. In contrast, our proposed Spoken-LLM methods perform slightly better than ParalinGPT on response text, with significantly superior performance on response style. Specifically, the Spoken-LLM-\textit{chunk} achieves 49.6 F1 score on response emotion with 62.1 F1 score on response speaking speed. \\
\textbf{Chunk vs. utterance-level embedding}: We compare the granularity of speech embedding on both ParalinGPT and Spoken-LLM methods. Results show that the use of chunk embedding achieves better performance on response style prediction. As for response text, the Spoken-LLM benefits significantly from chunk embedding while ParalinGPT performs similarly. In general, chunk embedding extracts richer style-related information than average-pooling embedding, which is more helpful in modeling response speech. 

\subsection{Subjective evaluation}
We perform the human listening evaluation to compare the generated samples of two methods. Specifically, we compare the proposed Spoken-LLM-\textit{chunk} with Text-LLM (text-only) and Text-LLM (cascaded) baseline. As shown in Figure \ref{fig:human_eval}, Spoken-LLM wins over the Text-LLM (text-only) method by a large margin, demonstrating that it is important to consider the speaking style information to respond properly. On the other hand, human listeners slightly prefer more on Text-LLM (cascaded) than Spoken-LLM. This result can be explained in two ways: 1) From the objective evaluation of response text, the performance of Spoken-LLM and Text-LLM (cascade) is similar, so the human listeners might not differentiate the content difference, and 2) for the response style, it is possible to respond with more than one response style but still sounds reasonably natural. Therefore, the current and response speaking style is not a one-to-one but a one-to-many relation. Future efforts should consider modeling with more than one response style for better performance and evaluation.

\section{Analyses}
\subsection{Same sentence in different speaking styles induce diverse responses}
Since the proposed StyleTalk evaluation set provides sets of the same dialogue context and current input content with two or three distinct speaking styles, we can analyze how diverse are the responses for each input speaking style. To measure the response text diversity, we adopt the self-BLEU~\citep{selfbleu} score to measure the diversity of each dialogue set. Precisely, we average the BLEU score of two response sentences given two speaking styles as the \textbf{dialogue set-level self-BLEU score}. The lower self-BLEU score indicates the generated text is more diverse according to different speaking styles. The results are shown in Table \ref{tab:self-bleu}. We observe that Spoken-LLM generates the most diverse response content compared to Text-LLM (cascaded) and ParalinGPT. In contrast, the Text-LLM (text-only) baseline generates the same content regardless of different speaking styles, yielding 100\% self-BLEU score.
\begin{table*}[]
\centering
\adjustbox{width=1\textwidth}{
\begin{tabular}{ll}
\toprule \toprule
\textbf{Dialogue context}                                                                                            & \begin{tabular}[c]{@{}l@{}}A : I've finally hit my goal of running five miles every day! \\ B : That's amazing, hard work really does pay off, doesn't it? \\ A : Absolutely, and I've never felt more alive!\end{tabular}                                                                                                                                                                                                                                                                                       \\ \hline
\textbf{Current text}                                                                                                & B:  I really need to get back into exercising too.                                                                                                                                                                                                                                                                                                                                                                                                                                                               \\ \hline \hline
\textbf{Not knowing current style}                                                                                   & \textbf{Text-LLM (text-only)}: Totally understand, let's start small together, maybe a walk or something?                                                                                                                                                                                                                                                                                                                                                                                                                 \\ \hline
\textbf{\begin{tabular}[c]{@{}l@{}}Current style in \\ \textless{}neutral, normal, quiet\textgreater{}\end{tabular}} & \begin{tabular}[c]{@{}l@{}}\textbf{Text-LLM (cascaded)}: \textless{}friendly, normal, normal\textgreater  Definitely! Let's find a day when we can go together; I could use the company.\\ \textbf{Spoken-LLM}: \textless{}cheerful, normal, normal\textgreater  That's the spirit! Let me know if you want any tips or company on your runs!\\ \textbf{Ground truth}: \textless{}friendly, normal, normal\textgreater  Hey, we could be workout buddies if you want, I can help you get started!\end{tabular}                              \\ \hline
\textbf{\begin{tabular}[c]{@{}l@{}}Current style in\\ \textless{}unfriendly, fast, loud\textgreater{}\end{tabular}}  & \begin{tabular}[c]{@{}l@{}}\textbf{Text-LLM (cascaded)}: \textless{}neutral, normal, normal\textgreater Hey, no pressure though, take things at your own pace, you know?\\ \textbf{Spoken-LLM}: \textless{}unfriendly, normal, normal\textgreater  Whoa, no need to stress out about it, we can start with something small if you want.\\ \textbf{Ground truth}: \textless{}unfriendly, normal, normal\textgreater  Whoa there, no need to get upset, maybe we can find a way to ease you into a routine? \end{tabular}  \\ \bottomrule \bottomrule                
\end{tabular}}
\caption{A qualitative example. The model outputs the response of speaker A's turn. }
\label{tab:qual}
\end{table*}

\subsection{Warmup pre-training and data quality}
Table \ref{tab:warmup} discusses different training strategies. Firstly, when we only utilize LLM-generated unfiltered data for training, despite the data amount being abundant compared to the train set, the performance of the response style is worse than the train set. Meanwhile, we observe that training on the unfiltered set can achieve better performance on response text, probably because the data amount of the train set is too small and prone to overfitting. We reveal that pre-training on the unfiltered set and fine-tuning on the train set (unfiltered$\rightarrow$train) can boost the performance significantly, which enables the model to learn the task and the common language usage first and then align the human standard with the train set.

\subsection{Qualitative example}
We show the qualitative example in Table \ref{tab:qual} with different models' outputs. This example shows that the Text-LLM (text-only) baseline predicts a more neutral sentiment response text, while the text-LLM (cascaded) and Spoken-LLM model generate text with a more aggressive and engaging tone, and the predicted response speaking styles are similar to the ground truth. More quantitative analyses are in Appendix \ref{appendix:transistion} (style transition) and \ref{appendix:diversity} (Diversity of current and responding style).


\section{Related works}
\noindent{\textbf{Speech-text Multimodal LLM}}: The progress in speech Self-supervised Learning (SSL)~\citep{ssl_review, superb, superb-sg, superb-prosody} and neural audio codec~\citep{zeghidour2021soundstream, defossez2022high, wu2023audiodec, yang2023hifi, kumar2023high} enable extracting discrete speech units, drawing attention to generative spoken language modeling. Specifically, the discrete speech units are treated as a special language for unit language modeling~\citep{audiolm, gslm, pgslm, dgslm, twist}, further enabling multiple speech processing tasks in single multimodal LLM~\citep{audiopalm, speechgpt, toward_joint, voxtlm, viola, speechllama, cosmic, spectron}. However, these works mostly leverage content information in speech, due to the speech unit clustering and the use of speech-transcript pairs for modality alignment. The multimodal LLM itself does not learn to model speaking style, and the models are mostly trained with single-turn utterances.

The other line of work aims for the universal speech and audio understanding model to have general audio perception and hearing ability, either leveraging off-the-shelf expert models~\citep{audiogpt, hugginggpt}, or with a single MM-LLM~\citep{qwen-audio, salmonn, ltu, ltu-as, pengi}. Those methods are mainly trained to perform comprehensive speech and audio understanding tasks and then fine-tuned for audio-based instruction-following data generated by off-the-shelf LLM like GPT-3.5. However, they are \textit{limited to only generating text responses without considering responding styles}, and the \textit{data quality from LLM is unknown}. In contrast, we focus on modeling speaking style in speech-to-speech conversation with a manual-filtered dataset. \\

\textbf{Speaking Style in Spoken Dialogue}: \label{sec:related}Speaking style is important for \textit{speech understanding} and \textit{response generation} in spoken dialogue. The understanding of speaking styles in spoken dialogue is crucial for extracting style attributes such as emotion, sentiment, sarcasm, and more. Representative corpora for studying speaking styles in spoken conversations include IEMOCAP~\citep{iemocap}, SEMAINE~\citep{semaine}, MUStARD~\citep{mustard}, Switchboard-sentiment~\citep{switchboard}, MELD~\citep{meld}, MEISD~\citep{meisd}, and MSP-improv~\citep{msp-improv}. These datasets are primarily constructed based on label annotations from real speech conversations (e.g., TV series) or acted spoken conversations.

Another research direction involves \textit{spoken conversation in the form of spoken question answering}. Datasets in this category include NMSQA~\citep{nmsqa}, SCQA~\citep{scqa}, OpenSAQA~\citep{ltu-as}, and E-chat200~\citep{echat}, where the data sample is presented as a tuple of (question, answer). Specifically, for style-related questions and answers, OpenSAQA employs GPT-3.5 to generate textual questions based on the speech content and metadata style information, while E-chat considers text with emotion labels as the question for GPT-3.5 to generate the responding answer as gold answers.

In all existing datasets, only one style is attached to the speech, and one corresponding response speech exists for each conversational context. Thus, prohibiting researchers from investigating the impact of different styles given the same context and the same words. Additionally, the SQA data in OpenSAQA and E-chat are fully generated by GPT-3.5 and not carefully checked by humans, resembling distillation and prompting of GPT-3.5, which is concerning whether the sample follows a human standard as spoken conversation. Our work provides the spoken dialogue data with the same context, the same current text with different speaking styles, and the corresponding response speech with human annotator filtering.

\section{Conclusion}
This paper focuses on enhancing LLM by modeling how the same sentence spoken with different speaking styles causes different responses in speech, in the spoken conversation scenario. Due to the absence of a suitable dataset, we first collect the speech-to-speech StyleTalk dataset that contains the same dialogue context the same sentence spoken in different styles, and the corresponding different response speech. Next, we propose Spoken-LLM, a two-stage multi-modal training framework to capture different speaking styles and respond properly. The proposed method yields better performance than the text and speech baseline on objective metrics and performs better than text-only LLM on subjective evaluation. We encourage the research community to use the released StyleTalk for joint speaking style and language modeling.

\section*{Limitation}
\textbf{Data scale}: The current StyleTalk training set consists of only around 2K samples, which may lead to training instability and overfitting. Utilizing a larger-scale dataset could alleviate these issues and eliminate the need for a pre-training stage on unfiltered LLM-generated data. \\ 
\textbf{Real speech with diverse and mixed styles}: The speech data in StyleTalk is synthesized from the Azure TTS system with style control. However, incorporating spontaneous speech with even more diverse styles is preferable. Moreover, the current one-hot emotion simplified the problem since speech emotion may by expressed multi-label distribution. \\
\textbf{Direct speech-to-speech modeling}: The Spoken-LLM generates predefined style attributes to feed into the expressive TTS system. Future work on directly modeling responding speech has the potential to eliminate the need for explicit style labels. \\
\textbf{Toward human-like spoken dialogue:} Real-world human communication includes backchannel, laughter, and turn-taking behaviors, which is beyond the turn-based spoken dialogue system~\citep{dgslm, chats}. Future endeavors to explore speaking style with those behaviors can make the spoken dialogue model closer to human conversation.

\section*{Acknowledgement}
We want to express our gratitude to the reviewers for their detailed feedback and actionable suggestions, which helped us strengthen our paper. Guan-Ting Lin is supported by a joint NTU GICE and Nvidia Ph.D. scholarship program. Additionally, we thank the National Center for High-performance Computing (NCHC) of the National Applied Research Laboratories (NARLabs) in Taiwan for providing computational and storage resources.

\bibliography{custom}

\begin{thebibliography}{55}
\expandafter\ifx\csname natexlab\endcsname\relax\def\natexlab#1{#1}\fi

\bibitem[{Baevski et~al.(2023)Baevski, Babu, Hsu, and Auli}]{data2vec2}
Alexei Baevski, Arun Babu, Wei-Ning Hsu, and Michael Auli. 2023.
\newblock Efficient self-supervised learning with contextualized target representations for vision, speech and language.
\newblock In \emph{International Conference on Machine Learning}, pages 1416--1429. PMLR.

\bibitem[{Banerjee and Lavie(2005)}]{meteor}
Satanjeev Banerjee and Alon Lavie. 2005.
\newblock Meteor: An automatic metric for mt evaluation with improved correlation with human judgments.
\newblock In \emph{Proceedings of the acl workshop on intrinsic and extrinsic evaluation measures for machine translation and/or summarization}, pages 65--72.

\bibitem[{Borsos et~al.(2023)Borsos, Marinier, Vincent, Kharitonov, Pietquin, Sharifi, Roblek, Teboul, Grangier, Tagliasacchi et~al.}]{audiolm}
Zal{\'a}n Borsos, Rapha{\"e}l Marinier, Damien Vincent, Eugene Kharitonov, Olivier Pietquin, Matt Sharifi, Dominik Roblek, Olivier Teboul, David Grangier, Marco Tagliasacchi, et~al. 2023.
\newblock Audiolm: a language modeling approach to audio generation.
\newblock \emph{IEEE/ACM Transactions on Audio, Speech, and Language Processing}.

\bibitem[{Busso et~al.(2008)Busso, Bulut, Lee, Kazemzadeh, Mower, Kim, Chang, Lee, and Narayanan}]{iemocap}
Carlos Busso, Murtaza Bulut, Chi-Chun Lee, Abe Kazemzadeh, Emily Mower, Samuel Kim, Jeannette~N Chang, Sungbok Lee, and Shrikanth~S Narayanan. 2008.
\newblock Iemocap: Interactive emotional dyadic motion capture database.
\newblock \emph{Language resources and evaluation}, 42:335--359.

\bibitem[{Busso et~al.(2016)Busso, Parthasarathy, Burmania, AbdelWahab, Sadoughi, and Provost}]{msp-improv}
Carlos Busso, Srinivas Parthasarathy, Alec Burmania, Mohammed AbdelWahab, Najmeh Sadoughi, and Emily~Mower Provost. 2016.
\newblock Msp-improv: An acted corpus of dyadic interactions to study emotion perception.
\newblock \emph{IEEE Transactions on Affective Computing}, 8(1):67--80.

\bibitem[{Castro et~al.(2019)Castro, Hazarika, P{\'e}rez-Rosas, Zimmermann, Mihalcea, and Poria}]{mustard}
Santiago Castro, Devamanyu Hazarika, Ver{\'o}nica P{\'e}rez-Rosas, Roger Zimmermann, Rada Mihalcea, and Soujanya Poria. 2019.
\newblock Towards multimodal sarcasm detection (an \_obviously\_ perfect paper).
\newblock In \emph{Proceedings of the 57th Annual Meeting of the Association for Computational Linguistics}, pages 4619--4629.

\bibitem[{Chen et~al.(2020)Chen, Lu, Xu, Cao, Zhang, and Fan}]{switchboard}
Eric Chen, Zhiyun Lu, Hao Xu, Liangliang Cao, Yu~Zhang, and James Fan. 2020.
\newblock A large scale speech sentiment corpus.
\newblock In \emph{Proc.\ LREC}, pages 6549--6555.

\bibitem[{Chou et~al.(2023)Chou, Chien, Hsu, Livescu, Babu, Conneau, Baevski, and Auli}]{toward_joint}
Ju-Chieh Chou, Chung-Ming Chien, Wei-Ning Hsu, Karen Livescu, Arun Babu, Alexis Conneau, Alexei Baevski, and Michael Auli. 2023.
\newblock Toward joint language modeling for speech units and text.
\newblock In \emph{Findings of the Association for Computational Linguistics: EMNLP 2023}, pages 6582--6593.

\bibitem[{Chu et~al.(2023)Chu, Xu, Zhou, Yang, Zhang, Yan, Zhou, and Zhou}]{qwen-audio}
Yunfei Chu, Jin Xu, Xiaohuan Zhou, Qian Yang, Shiliang Zhang, Zhijie Yan, Chang Zhou, and Jingren Zhou. 2023.
\newblock Qwen-audio: Advancing universal audio understanding via unified large-scale audio-language models.
\newblock \emph{arXiv preprint arXiv:2311.07919}.

\bibitem[{D{\'e}fossez et~al.(2022)D{\'e}fossez, Copet, Synnaeve, and Adi}]{defossez2022high}
Alexandre D{\'e}fossez, Jade Copet, Gabriel Synnaeve, and Yossi Adi. 2022.
\newblock High fidelity neural audio compression.
\newblock \emph{arXiv preprint arXiv:2210.13438}.

\bibitem[{Deshmukh et~al.(2023)Deshmukh, Elizalde, Singh, and Wang}]{pengi}
Soham Deshmukh, Benjamin Elizalde, Rita Singh, and Huaming Wang. 2023.
\newblock Pengi: An audio language model for audio tasks.
\newblock \emph{arXiv preprint arXiv:2305.11834}.

\bibitem[{Everson et~al.(2024)Everson, Gu, Yang, Shivakumar, Lin, Kolehmainen, Bulyko, Gandhe, Ghosh, Hamza et~al.}]{everson2024towards}
Kevin Everson, Yile Gu, Huck Yang, Prashanth~Gurunath Shivakumar, Guan-Ting Lin, Jari Kolehmainen, Ivan Bulyko, Ankur Gandhe, Shalini Ghosh, Wael Hamza, et~al. 2024.
\newblock Towards asr robust spoken language understanding through in-context learning with word confusion networks.
\newblock \emph{arXiv preprint arXiv:2401.02921}.

\bibitem[{Firdaus et~al.(2020)Firdaus, Chauhan, Ekbal, and Bhattacharyya}]{meisd}
Mauajama Firdaus, Hardik Chauhan, Asif Ekbal, and Pushpak Bhattacharyya. 2020.
\newblock Meisd: A multimodal multi-label emotion, intensity and sentiment dialogue dataset for emotion recognition and sentiment analysis in conversations.
\newblock In \emph{Proceedings of the 28th international conference on computational linguistics}, pages 4441--4453.

\bibitem[{Gong et~al.(2023{\natexlab{a}})Gong, Liu, Luo, Karlinsky, and Glass}]{ltu-as}
Yuan Gong, Alexander~H Liu, Hongyin Luo, Leonid Karlinsky, and James Glass. 2023{\natexlab{a}}.
\newblock Joint audio and speech understanding.
\newblock \emph{2023 IEEE Automatic Speech Recognition and Understanding Workshop (ASRU)}.

\bibitem[{Gong et~al.(2023{\natexlab{b}})Gong, Luo, Liu, Karlinsky, and Glass}]{ltu}
Yuan Gong, Hongyin Luo, Alexander~H Liu, Leonid Karlinsky, and James Glass. 2023{\natexlab{b}}.
\newblock Listen, think, and understand.
\newblock \emph{arXiv preprint arXiv:2305.10790}.

\bibitem[{Hassid et~al.(2023)Hassid, Remez, Nguyen, Gat, Conneau, Kreuk, Copet, D{\'e}fossez, Synnaeve, Dupoux, Schwartz, and Adi}]{twist}
Michael Hassid, Tal Remez, Tu~Anh Nguyen, Itai Gat, Alexis Conneau, Felix Kreuk, Jade Copet, Alexandre D{\'e}fossez, Gabriel Synnaeve, Emmanuel Dupoux, Roy Schwartz, and Yossi Adi. 2023.
\newblock \href {https://openreview.net/forum?id=UlHueVjAKr} {Textually pretrained speech language models}.
\newblock In \emph{Thirty-seventh Conference on Neural Information Processing Systems}.

\bibitem[{He and Garner(2023)}]{he23_interspeech}
Mutian He and Philip~N. Garner. 2023.
\newblock \href {https://doi.org/10.21437/Interspeech.2023-1799} {{Can ChatGPT Detect Intent? Evaluating Large Language Models for Spoken Language Understanding}}.
\newblock In \emph{Proc. INTERSPEECH 2023}, pages 1109--1113.

\bibitem[{Hu et~al.(2021)Hu, Wallis, Allen-Zhu, Li, Wang, Wang, Chen et~al.}]{lora}
Edward~J Hu, Phillip Wallis, Zeyuan Allen-Zhu, Yuanzhi Li, Shean Wang, Lu~Wang, Weizhu Chen, et~al. 2021.
\newblock Lora: Low-rank adaptation of large language models.
\newblock In \emph{International Conference on Learning Representations}.

\bibitem[{Huang et~al.(2023)Huang, Li, Yang, Shi, Chang, Ye, Wu, Hong, Huang, Liu et~al.}]{audiogpt}
Rongjie Huang, Mingze Li, Dongchao Yang, Jiatong Shi, Xuankai Chang, Zhenhui Ye, Yuning Wu, Zhiqing Hong, Jiawei Huang, Jinglin Liu, et~al. 2023.
\newblock Audiogpt: Understanding and generating speech, music, sound, and talking head.
\newblock \emph{arXiv preprint arXiv:2304.12995}.

\bibitem[{Kharitonov et~al.(2022)Kharitonov, Lee, Polyak, Adi, Copet, Lakhotia, Nguyen, Riviere, Mohamed, Dupoux et~al.}]{pgslm}
Eugene Kharitonov, Ann Lee, Adam Polyak, Yossi Adi, Jade Copet, Kushal Lakhotia, Tu-Anh Nguyen, Morgane Riviere, Abdelrahman Mohamed, Emmanuel Dupoux, et~al. 2022.
\newblock Text-free prosody-aware generative spoken language modeling.
\newblock In \emph{Proceedings of the 60th Annual Meeting of the Association for Computational Linguistics (Volume 1: Long Papers)}, pages 8666--8681.

\bibitem[{Kumar et~al.(2023)Kumar, Seetharaman, Luebs, Kumar, and Kumar}]{kumar2023high}
Rithesh Kumar, Prem Seetharaman, Alejandro Luebs, Ishaan Kumar, and Kundan Kumar. 2023.
\newblock \href {https://openreview.net/forum?id=qjnl1QUnFA} {High-fidelity audio compression with improved {RVQGAN}}.
\newblock In \emph{Thirty-seventh Conference on Neural Information Processing Systems}.

\bibitem[{Lakhotia et~al.(2021)Lakhotia, Kharitonov, Hsu, Adi, Polyak, Bolte, Nguyen, Copet, Baevski, Mohamed et~al.}]{gslm}
Kushal Lakhotia, Evgeny Kharitonov, Wei-Ning Hsu, Yossi Adi, Adam Polyak, Benjamin Bolte, Tu-Anh Nguyen, Jade Copet, Alexei Baevski, Abdelrahman Mohamed, et~al. 2021.
\newblock On generative spoken language modeling from raw audio.
\newblock \emph{Transactions of the Association for Computational Linguistics}, 9:1336--1354.

\bibitem[{Lin(2004)}]{rouge}
Chin-Yew Lin. 2004.
\newblock Rouge: A package for automatic evaluation of summaries.
\newblock In \emph{Text summarization branches out}, pages 74--81.

\bibitem[{Lin et~al.(2022)Lin, Chuang, Chung, wen Yang, Chen, Dong, Li, Mohamed, yi~Lee, and shan Lee}]{nmsqa}
Guan-Ting Lin, Yung-Sung Chuang, Ho-Lam Chung, Shu wen Yang, Hsuan-Jui Chen, Shuyan~Annie Dong, Shang-Wen Li, Abdelrahman Mohamed, Hung yi~Lee, and Lin shan Lee. 2022.
\newblock \href {https://doi.org/10.21437/Interspeech.2022-612} {{DUAL: Discrete Spoken Unit Adaptive Learning for Textless Spoken Question Answering}}.
\newblock In \emph{Proc. Interspeech 2022}, pages 5165--5169.

\bibitem[{Lin et~al.(2023{\natexlab{a}})Lin, Feng, Huang, Tseng, Lin, Li, Lee, and Ward}]{superb-prosody}
Guan-Ting Lin, Chi-Luen Feng, Wei-Ping Huang, Yuan Tseng, Tzu-Han Lin, Chen-An Li, Hung-yi Lee, and Nigel~G Ward. 2023{\natexlab{a}}.
\newblock On the utility of self-supervised models for prosody-related tasks.
\newblock In \emph{2022 IEEE Spoken Language Technology Workshop (SLT)}, pages 1104--1111. IEEE.

\bibitem[{Lin et~al.(2023{\natexlab{b}})Lin, Shivakumar, Gandhe, Yang, Gu, Ghosh, Stolcke, Lee, and Bulyko}]{paralingpt}
Guan-Ting Lin, Prashanth~Gurunath Shivakumar, Ankur Gandhe, Chao-Han~Huck Yang, Yile Gu, Shalini Ghosh, Andreas Stolcke, Hung-yi Lee, and Ivan Bulyko. 2023{\natexlab{b}}.
\newblock Paralinguistics-enhanced large language modeling of spoken dialogue.
\newblock \emph{arXiv preprint arXiv:2312.15316}.

\bibitem[{Ma et~al.(2023)Ma, Zheng, Ye, Li, Gao, Zhang, and Chen}]{emotion2vec}
Ziyang Ma, Zhisheng Zheng, Jiaxin Ye, Jinchao Li, Zhifu Gao, Shiliang Zhang, and Xie Chen. 2023.
\newblock emotion2vec: Self-supervised pre-training for speech emotion representation.
\newblock \emph{arXiv preprint arXiv:2312.15185}.

\bibitem[{Maiti et~al.(2023)Maiti, Peng, Choi, Jung, Chang, and Watanabe}]{voxtlm}
Soumi Maiti, Yifan Peng, Shukjae Choi, Jee-weon Jung, Xuankai Chang, and Shinji Watanabe. 2023.
\newblock Voxtlm: unified decoder-only models for consolidating speech recognition/synthesis and speech/text continuation tasks.
\newblock \emph{arXiv preprint arXiv:2309.07937}.

\bibitem[{McKeown et~al.(2010)McKeown, Valstar, Cowie, and Pantic}]{semaine}
Gary McKeown, Michel~F Valstar, Roderick Cowie, and Maja Pantic. 2010.
\newblock The semaine corpus of emotionally coloured character interactions.
\newblock In \emph{2010 IEEE International Conference on Multimedia and Expo}, pages 1079--1084. IEEE.

\bibitem[{Mitsui et~al.(2023)Mitsui, Hono, and Sawada}]{chats}
Kentaro Mitsui, Yukiya Hono, and Kei Sawada. 2023.
\newblock Towards human-like spoken dialogue generation between ai agents from written dialogue.
\newblock \emph{arXiv preprint arXiv:2310.01088}.

\bibitem[{Mohamed et~al.(2022)Mohamed, Lee, Borgholt, Havtorn, Edin, Igel, Kirchhoff, Li, Livescu, Maaløe, Sainath, and Watanabe}]{ssl_review}
Abdelrahman Mohamed, Hung-yi Lee, Lasse Borgholt, Jakob~D. Havtorn, Joakim Edin, Christian Igel, Katrin Kirchhoff, Shang-Wen Li, Karen Livescu, Lars Maaløe, Tara~N. Sainath, and Shinji Watanabe. 2022.
\newblock \href {https://doi.org/10.1109/JSTSP.2022.3207050} {Self-supervised speech representation learning: A review}.
\newblock \emph{IEEE Journal of Selected Topics in Signal Processing}, 16(6):1179--1210.

\bibitem[{Nachmani et~al.(2023)Nachmani, Levkovitch, Salazar, Asawaroengchai, Mariooryad, Skerry-Ryan, and Ramanovich}]{spectron}
Eliya Nachmani, Alon Levkovitch, Julian Salazar, Chulayutsh Asawaroengchai, Soroosh Mariooryad, RJ~Skerry-Ryan, and Michelle~Tadmor Ramanovich. 2023.
\newblock Lms with a voice: Spoken language modeling beyond speech tokens.
\newblock \emph{arXiv preprint arXiv:2305.15255}.

\bibitem[{Nguyen et~al.(2023)Nguyen, Kharitonov, Copet, Adi, Hsu, Elkahky, Tomasello, Algayres, Sagot, Mohamed et~al.}]{dgslm}
Tu~Anh Nguyen, Eugene Kharitonov, Jade Copet, Yossi Adi, Wei-Ning Hsu, Ali Elkahky, Paden Tomasello, Robin Algayres, Benoit Sagot, Abdelrahman Mohamed, et~al. 2023.
\newblock Generative spoken dialogue language modeling.
\newblock \emph{Transactions of the Association for Computational Linguistics}, 11:250--266.

\bibitem[{OpenAI(2023)}]{gpt4}
OpenAI. 2023.
\newblock \href {http://arxiv.org/abs/2303.08774} {Gpt-4 technical report}.

\bibitem[{Pan et~al.(2023)Pan, Wu, Gaur, Sivasankaran, Chen, Liu, and Li}]{cosmic}
Jing Pan, Jian Wu, Yashesh Gaur, Sunit Sivasankaran, Zhuo Chen, Shujie Liu, and Jinyu Li. 2023.
\newblock Cosmic: Data efficient instruction-tuning for speech in-context learning.
\newblock \emph{arXiv preprint arXiv:2311.02248}.

\bibitem[{Papineni et~al.(2002)Papineni, Roukos, Ward, and Zhu}]{bleu}
Kishore Papineni, Salim Roukos, Todd Ward, and Wei-Jing Zhu. 2002.
\newblock Bleu: a method for automatic evaluation of machine translation.
\newblock In \emph{Proceedings of the 40th annual meeting of the Association for Computational Linguistics}, pages 311--318.

\bibitem[{Poria et~al.(2019)Poria, Hazarika, Majumder, Naik, Cambria, and Mihalcea}]{meld}
Soujanya Poria, Devamanyu Hazarika, Navonil Majumder, Gautam Naik, Erik Cambria, and Rada Mihalcea. 2019.
\newblock Meld: A multimodal multi-party dataset for emotion recognition in conversations.
\newblock In \emph{Proceedings of the 57th Annual Meeting of the Association for Computational Linguistics}, pages 527--536.

\bibitem[{Radford et~al.(2023)Radford, Kim, Xu, Brockman, Mcleavey, and Sutskever}]{whisper}
Alec Radford, Jong~Wook Kim, Tao Xu, Greg Brockman, Christine Mcleavey, and Ilya Sutskever. 2023.
\newblock \href {https://proceedings.mlr.press/v202/radford23a.html} {Robust speech recognition via large-scale weak supervision}.
\newblock In \emph{Proceedings of the 40th International Conference on Machine Learning}, volume 202 of \emph{Proceedings of Machine Learning Research}, pages 28492--28518. PMLR.

\bibitem[{Rubenstein et~al.(2023)Rubenstein, Asawaroengchai, Nguyen, Bapna, Borsos, Quitry, Chen, Badawy, Han, Kharitonov et~al.}]{audiopalm}
Paul~K Rubenstein, Chulayuth Asawaroengchai, Duc~Dung Nguyen, Ankur Bapna, Zal{\'a}n Borsos, F{\'e}lix de~Chaumont Quitry, Peter Chen, Dalia~El Badawy, Wei Han, Eugene Kharitonov, et~al. 2023.
\newblock Audiopalm: A large language model that can speak and listen.
\newblock \emph{arXiv preprint arXiv:2306.12925}.

\bibitem[{Shen et~al.(2023)Shen, Song, Tan, Li, Lu, and Zhuang}]{hugginggpt}
Yongliang Shen, Kaitao Song, Xu~Tan, Dongsheng Li, Weiming Lu, and Yueting Zhuang. 2023.
\newblock Hugginggpt: Solving ai tasks with chatgpt and its friends in huggingface.
\newblock \emph{arXiv preprint arXiv:2303.17580}.

\bibitem[{Tang et~al.(2023)Tang, Yu, Sun, Chen, Tan, Li, Lu, Ma, and Zhang}]{salmonn}
Changli Tang, Wenyi Yu, Guangzhi Sun, Xianzhao Chen, Tian Tan, Wei Li, Lu~Lu, Zejun Ma, and Chao Zhang. 2023.
\newblock Salmonn: Towards generic hearing abilities for large language models.
\newblock \emph{arXiv preprint arXiv:2310.13289}.

\bibitem[{Touvron et~al.(2023)Touvron, Martin, Stone, Albert, Almahairi, Babaei, Bashlykov, Batra, Bhargava, Bhosale et~al.}]{llama2}
Hugo Touvron, Louis Martin, Kevin Stone, Peter Albert, Amjad Almahairi, Yasmine Babaei, Nikolay Bashlykov, Soumya Batra, Prajjwal Bhargava, Shruti Bhosale, et~al. 2023.
\newblock Llama 2: Open foundation and fine-tuned chat models.
\newblock \emph{arXiv preprint arXiv:2307.09288}.

\bibitem[{Tsai et~al.(2022)Tsai, Chang, Huang, Huang, Lakhotia, Yang, Dong, Liu, Lai, Shi et~al.}]{superb-sg}
Hsiang-Sheng Tsai, Heng-Jui Chang, Wen-Chin Huang, Zili Huang, Kushal Lakhotia, Shu-wen Yang, Shuyan Dong, Andy Liu, Cheng-I Lai, Jiatong Shi, et~al. 2022.
\newblock Superb-sg: Enhanced speech processing universal performance benchmark for semantic and generative capabilities.
\newblock In \emph{Proceedings of the 60th Annual Meeting of the Association for Computational Linguistics (Volume 1: Long Papers)}, pages 8479--8492.

\bibitem[{Wang et~al.(2023)Wang, Zhou, Zhang, Wu, Liu, Gaur, Chen, Li, and Wei}]{viola}
Tianrui Wang, Long Zhou, Ziqiang Zhang, Yu~Wu, Shujie Liu, Yashesh Gaur, Zhuo Chen, Jinyu Li, and Furu Wei. 2023.
\newblock Viola: Unified codec language models for speech recognition, synthesis, and translation.
\newblock \emph{arXiv preprint arXiv:2305.16107}.

\bibitem[{Wei et~al.(2022)Wei, Tay, Bommasani, Raffel, Zoph, Borgeaud, Yogatama, Bosma, Zhou, Metzler et~al.}]{llm1}
Jason Wei, Yi~Tay, Rishi Bommasani, Colin Raffel, Barret Zoph, Sebastian Borgeaud, Dani Yogatama, Maarten Bosma, Denny Zhou, Donald Metzler, et~al. 2022.
\newblock Emergent abilities of large language models.
\newblock \emph{arXiv preprint arXiv:2206.07682}.

\bibitem[{Wu et~al.(2023{\natexlab{a}})Wu, Gaur, Chen, Zhou, Zhu, Wang, Li, Liu, Ren, Liu et~al.}]{speechllama}
Jian Wu, Yashesh Gaur, Zhuo Chen, Long Zhou, Yimeng Zhu, Tianrui Wang, Jinyu Li, Shujie Liu, Bo~Ren, Linquan Liu, et~al. 2023{\natexlab{a}}.
\newblock On decoder-only architecture for speech-to-text and large language model integration.
\newblock In \emph{2023 IEEE Automatic Speech Recognition and Understanding Workshop (ASRU)}, pages 1--8. IEEE.

\bibitem[{Wu et~al.(2023{\natexlab{b}})Wu, Gebru, Markovi{\'c}, and Richard}]{wu2023audiodec}
Yi-Chiao Wu, Israel~D Gebru, Dejan Markovi{\'c}, and Alexander Richard. 2023{\natexlab{b}}.
\newblock Audiodec: An open-source streaming high-fidelity neural audio codec.
\newblock In \emph{ICASSP 2023-2023 IEEE International Conference on Acoustics, Speech and Signal Processing (ICASSP)}, pages 1--5. IEEE.

\bibitem[{Xue et~al.(2023)Xue, Liang, Mu, Zhang, Chen, and Xie}]{echat}
Hongfei Xue, Yuhao Liang, Bingshen Mu, Shiliang Zhang, Qian Chen, and Lei Xie. 2023.
\newblock E-chat: Emotion-sensitive spoken dialogue system with large language models.
\newblock \emph{arXiv preprint arXiv:2401.00475}.

\bibitem[{Yang et~al.(2023)Yang, Liu, Huang, Tian, Weng, and Zou}]{yang2023hifi}
Dongchao Yang, Songxiang Liu, Rongjie Huang, Jinchuan Tian, Chao Weng, and Yuexian Zou. 2023.
\newblock Hifi-codec: Group-residual vector quantization for high fidelity audio codec.
\newblock \emph{arXiv preprint arXiv:2305.02765}.

\bibitem[{Yang et~al.(2021)Yang, Chi, Chuang, Lai, Lakhotia, Lin, Liu, Shi, Chang, Lin, Huang, Tseng, tik Lee, Liu, Huang, Dong, Li, Watanabe, Mohamed, and yi~Lee}]{superb}
Shu-Wen Yang, Po-Han Chi, Yung-Sung Chuang, Cheng-I~Jeff Lai, Kushal Lakhotia, Yist~Y. Lin, Andy~T. Liu, Jiatong Shi, Xuankai Chang, Guan-Ting Lin, Tzu-Hsien Huang, Wei-Cheng Tseng, Ko~tik Lee, Da-Rong Liu, Zili Huang, Shuyan Dong, Shang-Wen Li, Shinji Watanabe, Abdelrahman Mohamed, and Hung yi~Lee. 2021.
\newblock \href {https://doi.org/10.21437/Interspeech.2021-1775} {{SUPERB: Speech Processing Universal PERformance Benchmark}}.
\newblock In \emph{Proc. Interspeech 2021}, pages 1194--1198.

\bibitem[{You et~al.(2022)You, Chen, Liu, Ge, Wu, and Zou}]{scqa}
Chenyu You, Nuo Chen, Fenglin Liu, Shen Ge, Xian Wu, and Yuexian Zou. 2022.
\newblock End-to-end spoken conversational question answering: Task, dataset and model.
\newblock In \emph{Findings of the Association for Computational Linguistics: NAACL 2022}, pages 1219--1232.

\bibitem[{Zeghidour et~al.(2021)Zeghidour, Luebs, Omran, Skoglund, and Tagliasacchi}]{zeghidour2021soundstream}
Neil Zeghidour, Alejandro Luebs, Ahmed Omran, Jan Skoglund, and Marco Tagliasacchi. 2021.
\newblock Soundstream: An end-to-end neural audio codec.
\newblock \emph{IEEE/ACM Transactions on Audio, Speech, and Language Processing}, 30:495--507.

\bibitem[{Zhang et~al.(2023)Zhang, Li, Zhang, Zhan, Wang, Zhou, and Qiu}]{speechgpt}
Dong Zhang, Shimin Li, Xin Zhang, Jun Zhan, Pengyu Wang, Yaqian Zhou, and Xipeng Qiu. 2023.
\newblock \href {https://doi.org/10.18653/v1/2023.findings-emnlp.1055} {{S}peech{GPT}: Empowering large language models with intrinsic cross-modal conversational abilities}.
\newblock In \emph{Findings of the Association for Computational Linguistics: EMNLP 2023}, pages 15757--15773, Singapore. Association for Computational Linguistics.

\bibitem[{Zhang et~al.(2019)Zhang, Kishore, Wu, Weinberger, and Artzi}]{bertscore}
Tianyi Zhang, Varsha Kishore, Felix Wu, Kilian~Q Weinberger, and Yoav Artzi. 2019.
\newblock Bertscore: Evaluating text generation with bert.
\newblock In \emph{International Conference on Learning Representations}.

\bibitem[{Zhu et~al.(2018)Zhu, Lu, Zheng, Guo, Zhang, Wang, and Yu}]{selfbleu}
Yaoming Zhu, Sidi Lu, Lei Zheng, Jiaxian Guo, Weinan Zhang, Jun Wang, and Yong Yu. 2018.
\newblock Texygen: A benchmarking platform for text generation models.
\newblock In \emph{The 41st international ACM SIGIR conference on research \& development in information retrieval}, pages 1097--1100.

\end{thebibliography}
\bibliographystyle{acl_natbib}

\appendix

\section*{Appendix}
\label{sec:appendix}
\begin{table*}[]
\adjustbox{width=1\textwidth}{
\begin{tabular}{l|cccc|ccc}
\toprule
\multirow{2}{*}{\textbf{Method}}           & \multicolumn{4}{c}{\textbf{Response text}}                 & \multicolumn{3}{|c}{\textbf{Response style}} \\
                                   & \textbf{BLEU} & \textbf{ROUGE$_l$} & \textbf{METEOR} & \textbf{BERT$_{f1}$}  & \textbf{F1$_{emotion}$} & \textbf{F1$_{speed}$} & \textbf{F1$_{volume}$}\\ \hline
                                  
Text-LLM (cascaded)                     &   3.1 (-0.1)   &   16.9 (-0.4)   &     18.5 (-0.6)        &  75.9 (-0.1)    &     37.0 (-0.5)   &   52.5 (-0.4)  &   63.7 (-2.1)        \\ 
Spoken-LLM-\textit{chunk}  &  3.3 (-0.7)    &  17.1 (-0.9)     &   19.0 (-0.4) &  75.9 (-0.4)   &    47.5 (-2.1)      &     60.3 (-1.8)     & 57.4 (-3.7) \\ 
\bottomrule        
\end{tabular}
}
\caption{The results of using whisper base ASR model's prediction as input current text on text-LLM (cascaded) and Spoken-LLM-\textit{chunk}.}
\label{tab:asr}
\end{table*}

\section{Details of human annotators filtering}
\label{sec:manual}
\begin{figure*}{}
    \centering
    \begin{mdframed}
\includegraphics[width=1\linewidth]{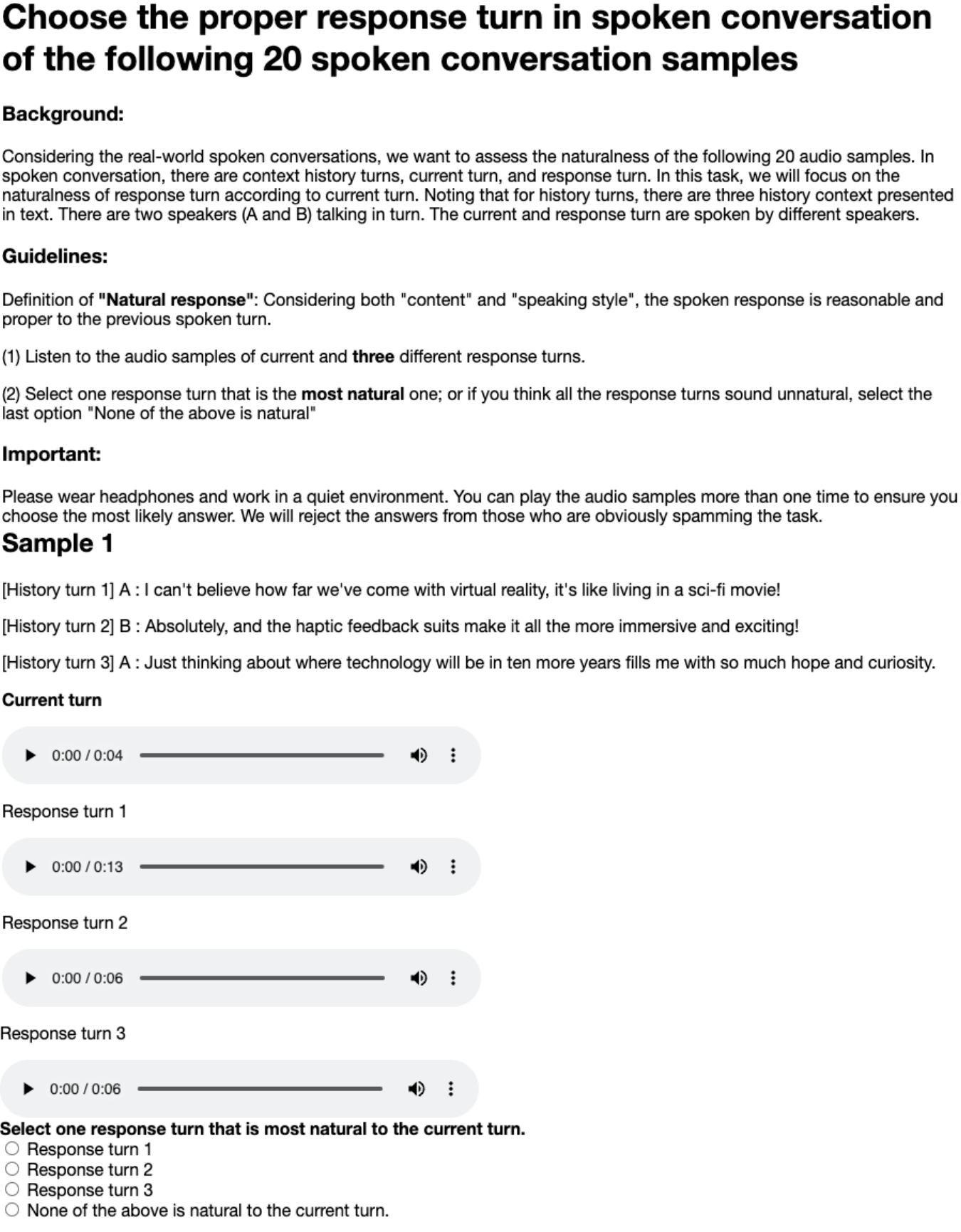}
    \end{mdframed}
    \caption{Human annotators filtering template.}
    \label{fig:data-annotate-template}
\end{figure*}

We assign three listeners for each test. All listeners are based in the United States with HIT approval rate higher than 95\%, given that the corpus is in American English. Only the pairs that receive a majority vote and do not have anyone choosing "None of the above is natural" are retained in our corpus. Each test contains 20 samples for evaluation. The example of the annotation interface is shown in Figure \ref{fig:data-annotate-template}. We pay the annotators 3 USD for each test. On average, based on the time of playing audio (if played twice for each sample) and reading the content, it takes 10 minutes on one test, so the hourly wage is around 18 USD.

\section{Implementation details}
The model is trained using a two-stage approach with distinct learning rates. The learning rate is 1e-3 and 2e-4 during the 1-stage and 2-stage, respectively. The batch size is 128, and LoRA (r=8) is utilized for efficient fine-tuning of the LLM. To facilitate stable training, a warmup learning rate strategy is with 100 initial steps then linear decay. We use 10\% of the training samples as the validation data to assess model performance during training. Model checkpoint is selected based on the validation set performance. In the inference stage, a temperature of 0.7 was applied to control the randomness of generated outputs, and top-p sampling with a probability threshold of 0.95 was used. The number of trainable parameters is 7.8M (0.11\% for total parameters). All experiments are run with a single A40 48G GPU.

\section{ASR prediction as input}
\label{sec:asr}

In this particular setup, we use the Whisper base ASR~\citep{whisper} model to transcribe the current speech into text, which is then input into the trained model for inference. The Word Error Rate (WER) on the current turn speech within the evaluation set is 3.21\%. In this setup, we test with the text-LLM (cascaded) and Spoken-LLM-\textit{chunk} models. In Table \ref{tab:asr}, compared to using the ground truth transcripts, we observe slight performance degradation in response text and style for both the Spoken-LLM-\textit{chunk} and text-LLM (cascaded). It's important to note that addressing ASR error propagation on LLM is beyond the scope of this paper. However, several previous efforts have delved into investigating methods to mitigate such issues~\citep{he23_interspeech, everson2024towards}, which may be one of the further directions especially when the more expressive and spontaneous speech as current input speech.

\section{Style transition}
\label{appendix:transistion}
In this section, we delve into an analysis of the correlation between input and output emotions. While the dataset comprises diverse samples with varying dialogue contexts and inputs, human responses exhibit discernible patterns associated with specific styles. Notably, individuals are inclined to respond with particular styles given a certain current style, and conversely, they are less likely to adopt certain styles in their responses. For instance, in cases where the input style is cheerful, the corresponding response style is more inclined towards positivity, such as cheerful and friendly emotions, as opposed to styles such as unfriendly or sad.

\begin{figure}[t]{}
\centering
\includegraphics[width=1\linewidth]{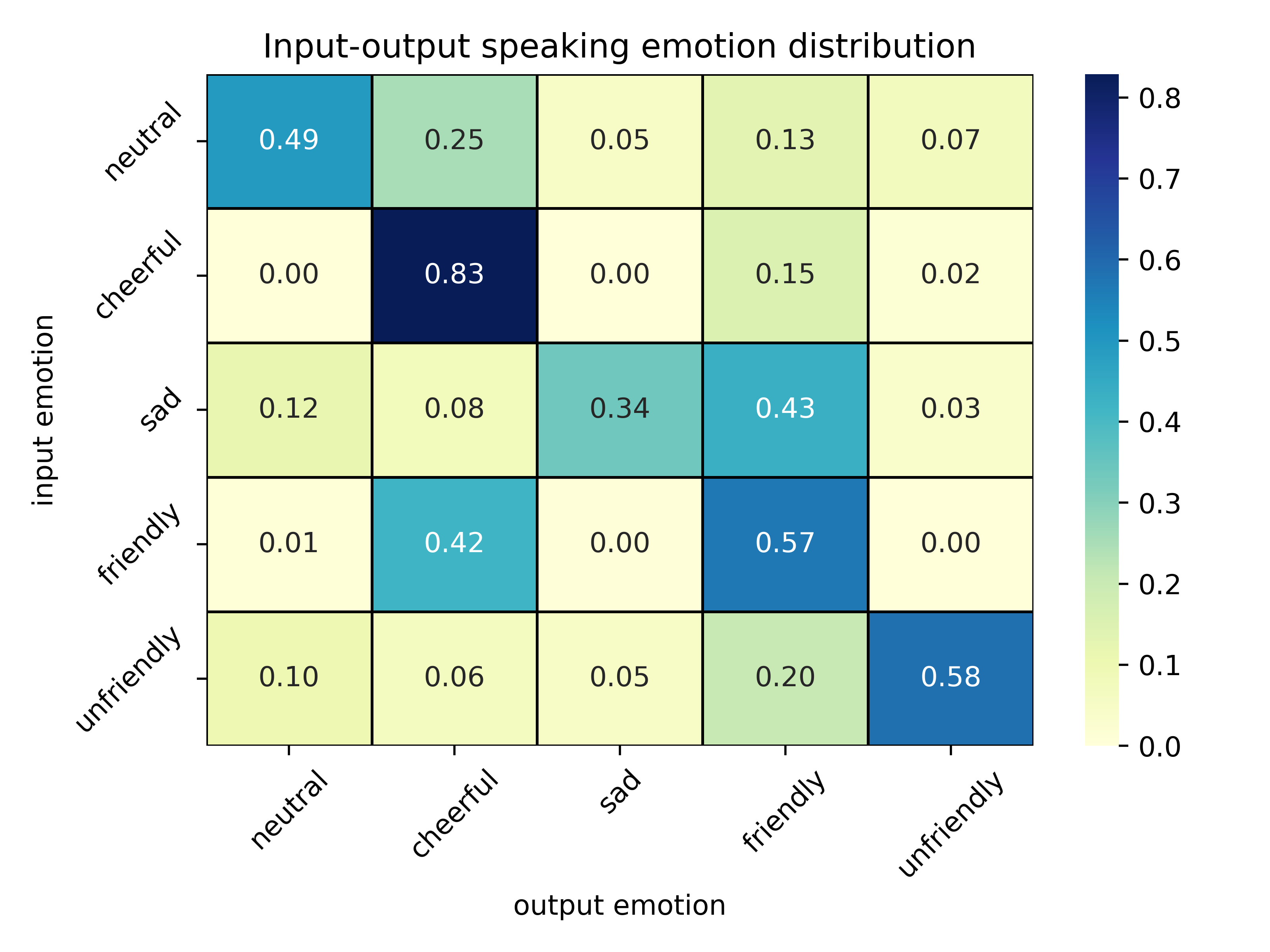}
    \caption{The output emotion distribution given input emotion. Each row is the probability distribution for an input-output pair.}
    \label{fig:distribution}
\end{figure}
In Figure \ref{fig:distribution}, we present a visual representation of the output style distribution corresponding to different input styles. The visualization reveals that for each input style, certain response styles are markedly more prevalent than others, underscoring the nuanced relationship between input and output emotions.

\section{Diversity of current and responding styles}
\label{appendix:diversity}
In exploring human responses across varied styles, individuals may employ more assertive or passive speaking approaches, resulting in potential differences in content. In this context, we delve into an examination of how the interplay between input and output styles influences response diversity. Specifically, we seek to determine whether the transition between styles in certain scenarios leads to responses characterized by increased diversity or a tendency to adopt simpler, more predictable patterns. This investigation sheds light on the intricate dynamics of style transitions and their impact on the richness and complexity of response text. 

\begin{figure}[t]{}
\centering
\includegraphics[width=1\linewidth]{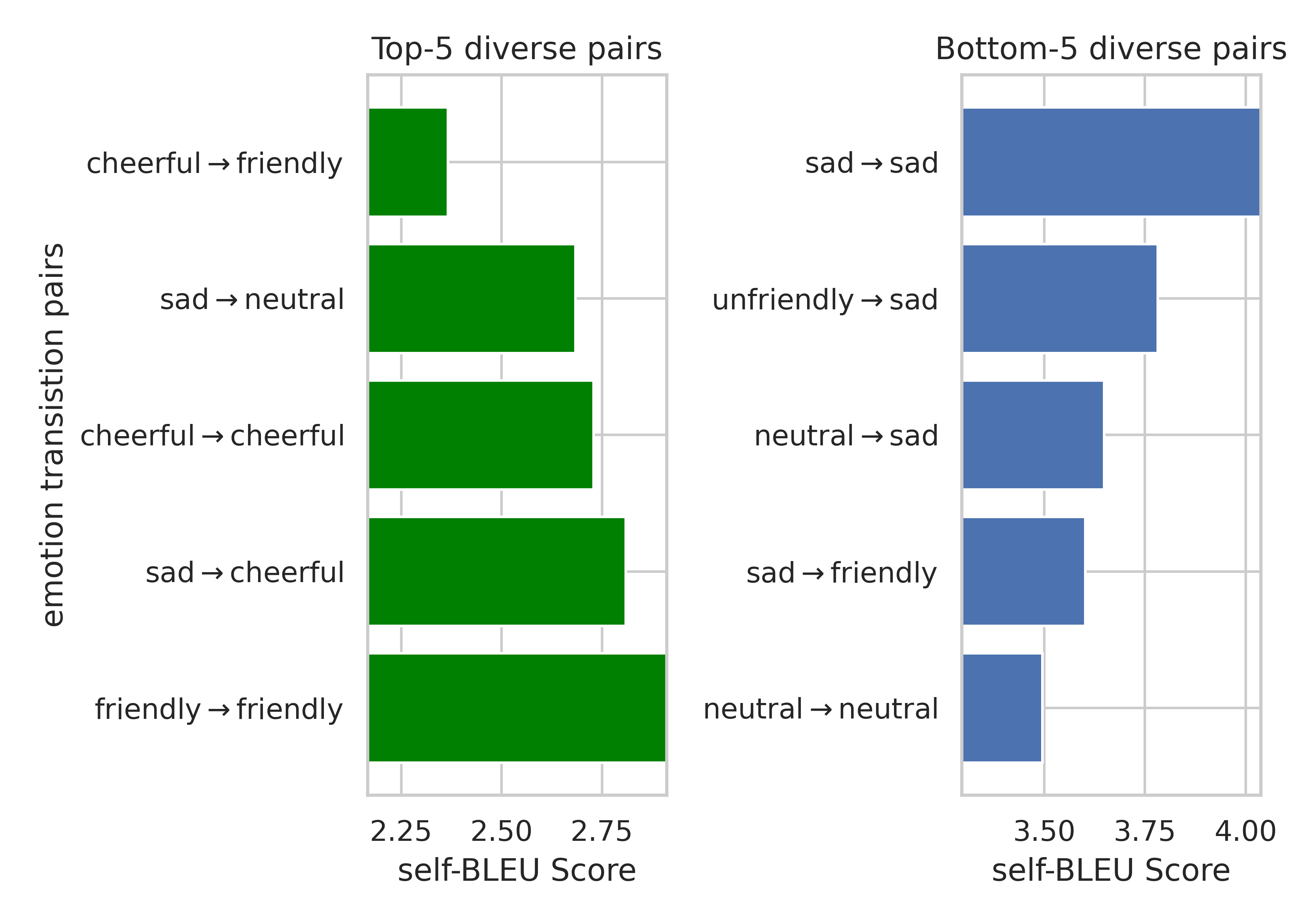}
    \caption{Top-5 and Bottom-5 diverse pairs in the train and evaluation set. The self-BLEU is normalized for each style transition pair to make a fair comparison. The pairs with fewer than 5 pairs are removed. The lower the self-BLEU score, the more diverse the lexical response given different dialogue contexts and input. }
    \label{fig:self-bleu}
\end{figure}

In Figure \ref{fig:self-bleu}, we present the result of the top-5 and bottom-5 diverse pairs, organized according to their self-BLEU scores. The self-BLEU score represents the average BLEU score for each style transition pair, with lower scores indicating greater diversity in the responses. Notably, we observe that the top-5 diverse pairs frequently involve responses characterized by positive and excitement styles such as cheerfulness. Conversely, the bottom-5 non-diverse pairs are associated with empathy, particularly when the input style is sad. This analysis provides insights into the response diversity across various style transition scenarios, emphasizing notable patterns in the use of distinct emotional styles.

\begin{table}[]
\begin{tabular}{llll}
\toprule
                  & \textbf{eval} & \textbf{train} & \textbf{unfiltered} \\ \midrule
\# dialogue set 1 & 16   & 1,770 & 0          \\
\# dialogue set 2 & 445  & 108   & 859        \\
\# dialogue set 3 & 25   & 0     & 4,918      \\
\# sample         & 981  & 1,986 & 16,472   \\ \bottomrule
\end{tabular}
\caption{Data statistics of StyleTalk. The \# dialogue set 1, 2, and 3 mean the amount of different speaking styles for the current speech. \# sample is the number of current and response speech pairs.}
\label{tab:data-statistic}
\end{table}

\section{Instruction}
\label{sec:instruction}
$I_1$: Instruction: Classify speaking style of speech. The speaking style is represented in (emotion, speed, volume).\\
$I_2$: Instruction: Generate human-like response given context. speaking style is represented in (emotion, speed, volume).

\section{Prompting GPT-4 for Data Generation}
\begin{figure*}{}
    \centering
    \begin{mdframed}
\includegraphics[width=1\linewidth]{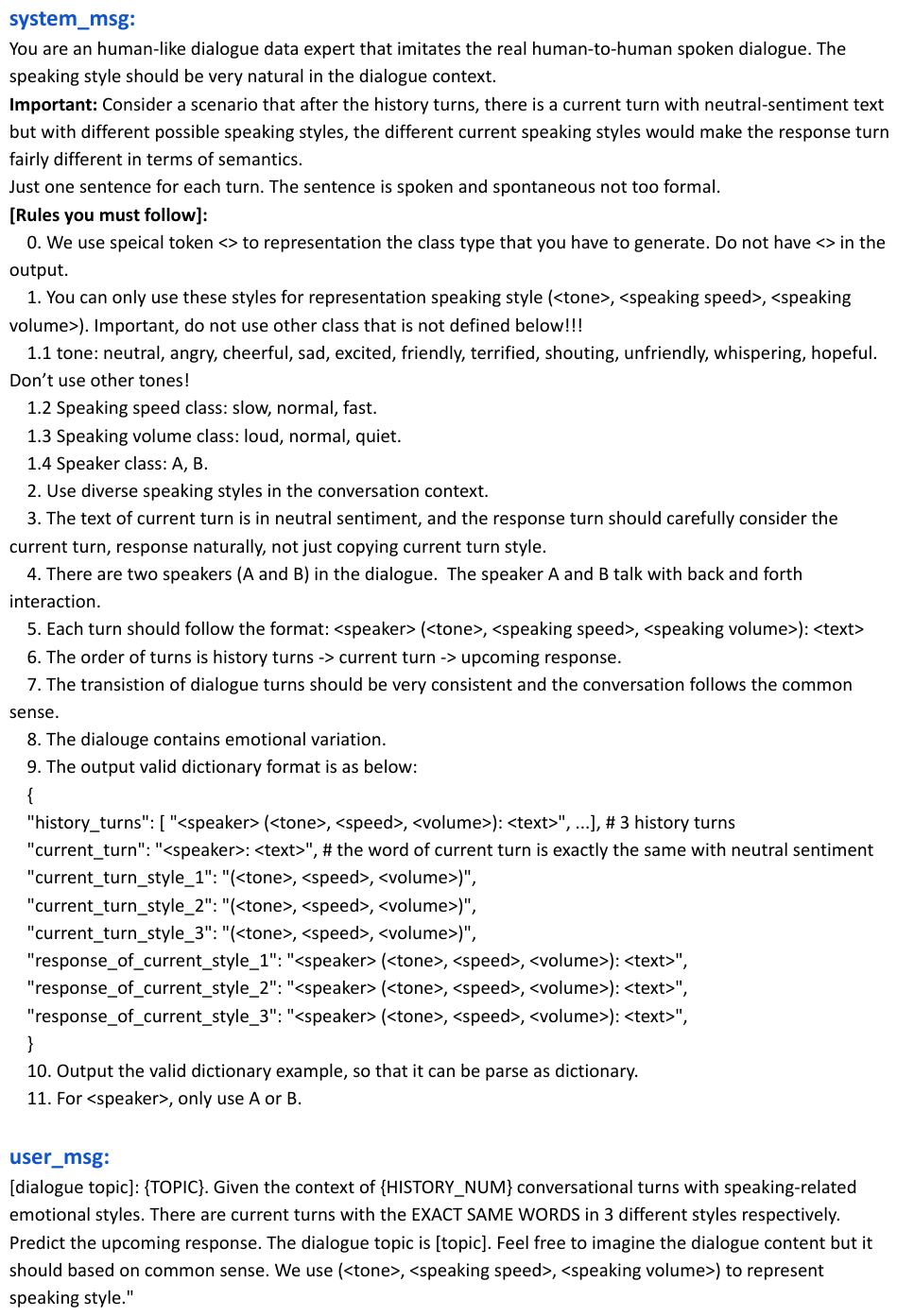}
    \end{mdframed}
    \caption{Prompt template. \{TOPIC\} and \{HISTORY\_NUM\} are variables. The system message and user message are sent to GPT-4 (\texttt{gpt-4-1106-preview}) API.}
    \label{fig:prompting}
\end{figure*}

We utilize \texttt{gpt-4-1106-preview} and the prompt template in Figure \ref{fig:prompting}.

\begin{figure*}{}
    \centering
    \begin{mdframed}
\includegraphics[width=1\linewidth]{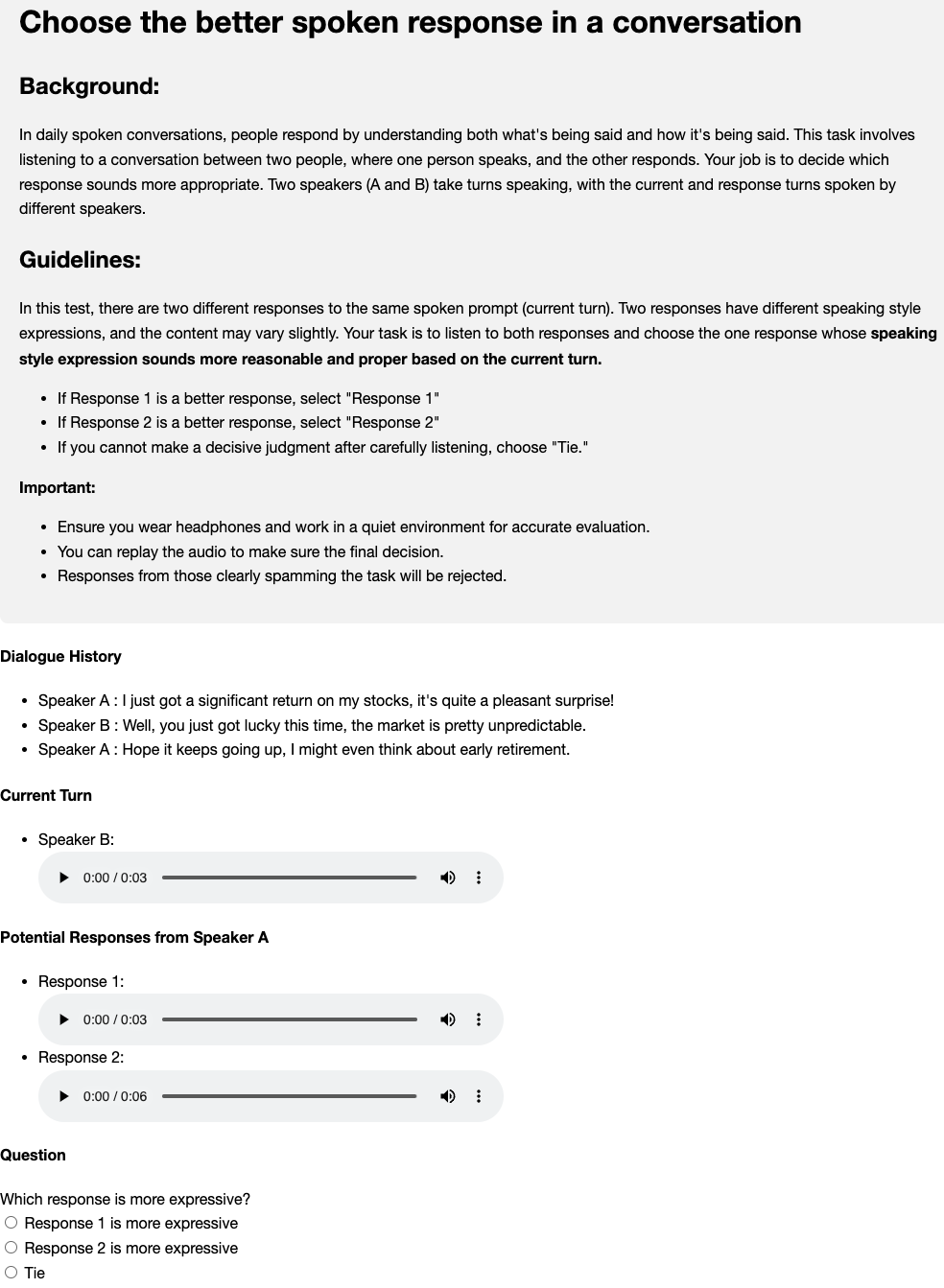}
    \end{mdframed}
    \caption{Subjective evaluation template.}
    \label{fig:subject-eval-template}
\end{figure*}

\section{Subjective evaluation}
\label{sec:subjective-eval}
Each test contains 10 samples for evaluation. The example of subjective evaluation interface is shown in Figure \ref{fig:subject-eval-template}. We pay the annotators 3 USD for each test. On average, based on the time of listening to audio (if played three times for each sample) and reading the content, it takes 10 minutes on one test. The hourly wage is around 18 USD.

\section{Dataset license}
We released the StyleTalk dataset under the MIT license.

\end{document}